\documentclass[letterpaper, 10 pt, conference]{ieeeconf}  

\IEEEoverridecommandlockouts                              

\usepackage{xcolor}

\usepackage{graphicx}

\usepackage{gensymb}
\usepackage[hyphens]{url}
\usepackage{multirow}
\usepackage{multicol}
\usepackage{tabularx}
\usepackage{dsfont}
\usepackage{url}
\usepackage{color}
\usepackage{subcaption}
\usepackage{caption}
\usepackage{booktabs}
\usepackage{dcolumn}
\usepackage{amssymb}
\usepackage{hyperref}
\usepackage{amsmath}
\usepackage{colortbl}
\usepackage{tikz}
\colorlet{lightgray}{gray!20}

\usepackage{physics}
\usepackage{graphicx}
\usepackage{float}
\usepackage{subfloat}
\usepackage{longtable}
\usepackage{threeparttable}
\usepackage{xcolor}
\usepackage{diagbox}
\usepackage{cite}

\usepackage{color}
\hypersetup{
    colorlinks=false,
    linkcolor=blue,
    filecolor=magenta,      
    urlcolor=cyan,
    pdftitle={Overleaf Example},
    pdfpagemode=FullScreen,
    }

\usepackage{booktabs}
\usepackage{multirow}

 \renewcommand{\paragraph}[1]{
    \vspace{2mm}
     \noindent\textbf{#1} 
 }

\definecolor{amber(sae/ece)}{rgb}{1.0, 0.49, 0.0}

\title{\LARGE \bf
Breaking Data Silos: Cross-Domain Learning for Multi-Agent Perception from Independent Private Sources}

\author{Jinlong Li$^{1}$, Baolu Li$^{1}$, Xinyu Liu$^{1}$,  Runsheng Xu$^{2}$, Jiaqi Ma$^{2}$, Hongkai Yu$^{1*}$ 
\thanks{$^{1}$Cleveland State University,  Cleveland Vision $\&$ AI Lab.   $^{2}$University of California, Los Angeles,  UCLA Mobility Lab. }
\thanks{*Corresponding Author: h.yu19@csuohio.edu}
}

\begin{document}

 \maketitle
\thispagestyle{empty}
\pagestyle{empty}

\begin{abstract}


The diverse agents in multi-agent perception systems may be from different companies. Each company might use the identical classic neural network architecture based encoder for feature extraction. However, the data source to train the various agents is independent and private in each company, leading to the Distribution Gap of different private data for training distinct agents in multi-agent perception system. The data silos by the above Distribution Gap could result in a significant performance decline in multi-agent perception. In this paper, we thoroughly examine the impact of the distribution gap on existing multi-agent perception systems. To break the data silos, we introduce the Feature Distribution-aware Aggregation (FDA) framework for cross-domain learning to mitigate the above Distribution Gap in multi-agent perception. FDA comprises two key components: Learnable Feature Compensation Module and   Distribution-aware Statistical Consistency Module, both aimed at enhancing intermediate features to minimize the distribution gap among multi-agent features. Intensive experiments on the public OPV2V and V2XSet datasets underscore FDA's effectiveness in point cloud-based 3D object detection, presenting it as an invaluable augmentation to existing multi-agent perception systems. The code is available at https://github.com/jinlong17/BDS-V2V. 
\end{abstract}

\section{Introduction}

%

Although advancements in deep learning have greatly enhanced single-agent perception, the challenges of long-range detection,  occlusion issues, and field of view still persist~\cite{chen2019f,xu2023v2v4real,tang2023collaborative}. Multi-agent perception systems offer a solution by leveraging Vehicle-to-Everything (V2X) communication technology, allowing multiple nearby agents to share visual information, including detection results, raw sensor data, and intermediate  features~\cite{wang2020v2vnet,xu2022opv2v,huwhere2comm,xu2023bridging}.

\begin{figure}[htb] 
    \begin{centering}
        \includegraphics[width=1
        \columnwidth]{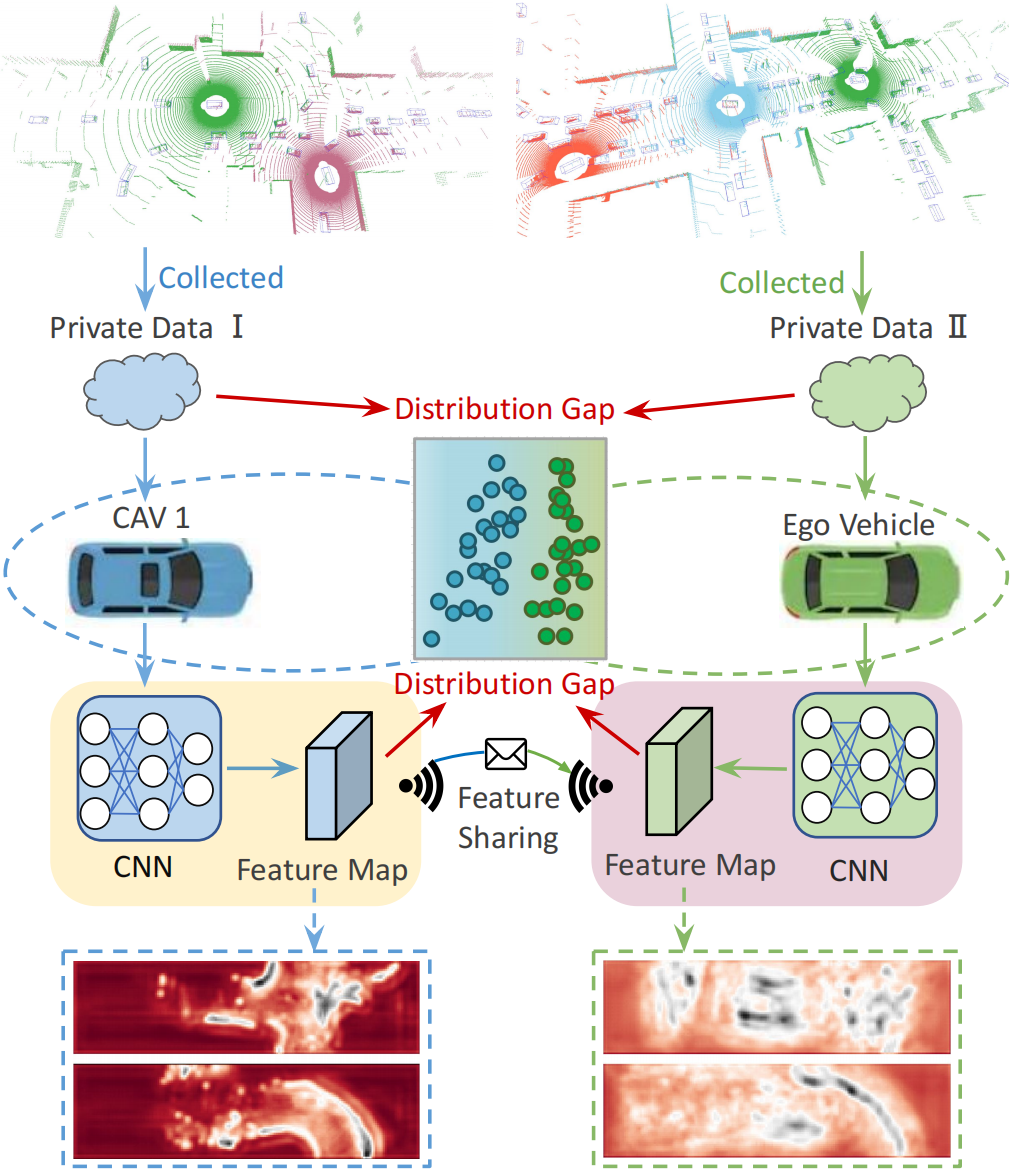}
    \par\end{centering}
    \caption{\textbf{Illustration of the Distribution Gap of different independent private data for training distinct agents in multi-agent perception}. Here we use V2X cooperative perception in autonomous driving as an example.}
    \label{fig:distribution_gap}
\end{figure}

%

Existing multi-agent perception methods often operate under a strong assumption: all agents are trained using identical training data. However, this assumption might be not true. The diverse agents in multi-agent perception systems may be from different companies. Some companies whose perception systems are established on the same popular  open platform might use the identical classic neural network architecture-based encoder for feature extraction. However, the data source to train the various agents is independent and private in each company as depicted in Fig.~\ref{fig:distribution_gap}, leading to the Distribution Gap of different private data for training distinct agents in multi-agent perception systems. As an example of Distribution Gaps, let us compare the two widely-used large-scale simulated V2X cooperative perception datasets, namely OPV2V~\cite{xu2022opv2v} and V2XSet~\cite{xu2022v2x}. We find key differences in the following two perspectives.

\begin{figure*}[htb] 
    \begin{centering}
        \includegraphics[width=1\textwidth]{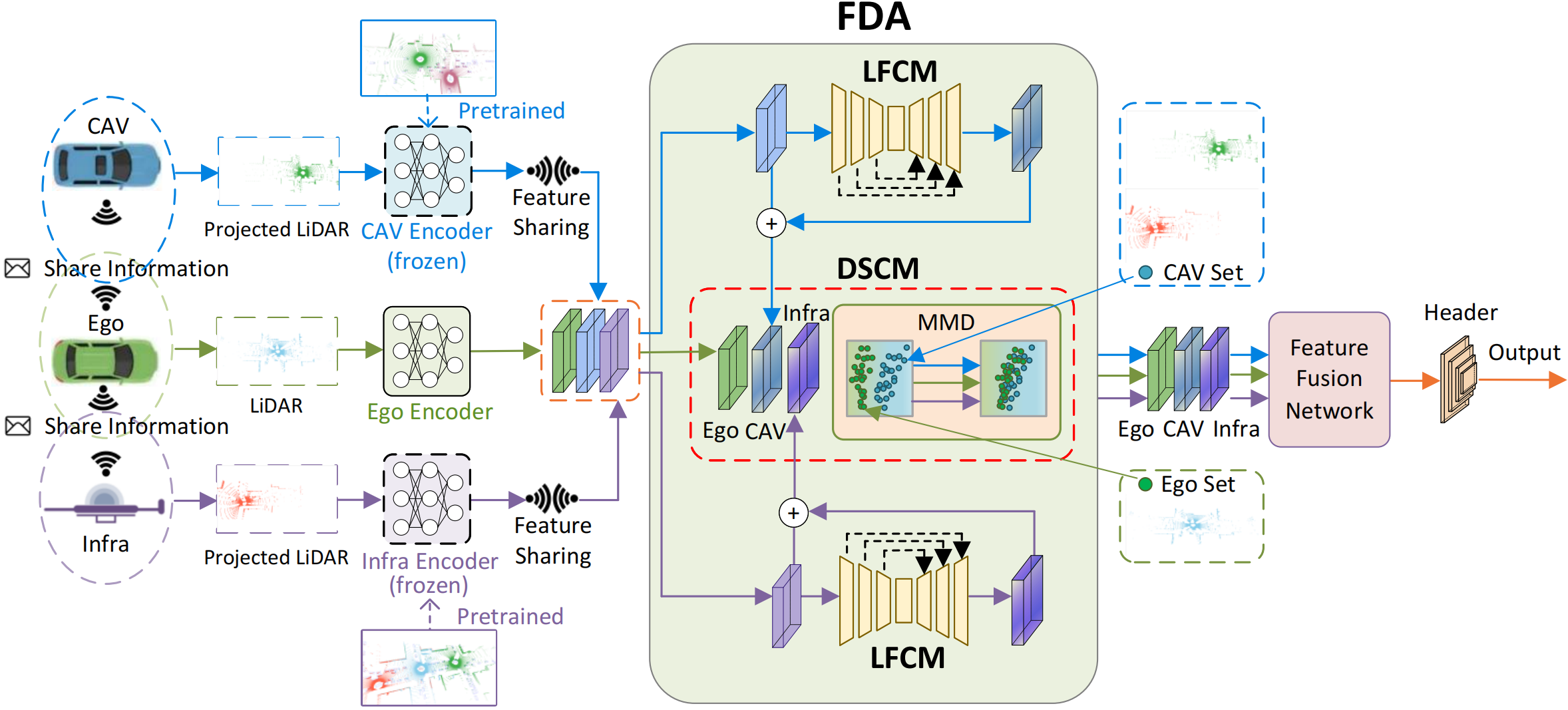}
    \par\end{centering}
    \caption{\textbf{Architecture of the proposed Feature Distribution-aware Aggregation (FDA)  framework.} FDA leverages LFCM to generate residual compensation map to enhance other-agent  features, then utilizes DSCM to mitigate Distribution Gaps.}
    \label{fig:FDA}
\end{figure*}

\begin{itemize}
\item \textbf{Agent Number and Type:} The presence of diverse agent numbers within the same scenario can yield a different pool of shared visual information. Variations in the quantity and types of agents, \textit{e.g.}, Connected and Automated Vehicles (CAVs), and Infrastructures can also contribute to disparities in the sharing of intermediate features.

\item \textbf{Scene:} The density of LiDAR point cloud data can vary significantly based on the size and shape of the roadway environment in which the agents operate, leading to fluctuations in feature sharing.
\end{itemize}

The data silos by the above Distribution Gap could result in a significant performance decline in multi-agent perception. A naive solution for solving this problem might be using federated learning-based methods. Federated learning represents a privacy-preserving collaborative machine learning paradigm~\cite{mcmahan2017communication,yang2019federated}, designed to facilitate the collaborative training of a shared global model among multiple clients by exchanging information pertaining to model parameters. However, in the autonomous driving industry, \textit{the data collection and annotation are extremely expensive and with business secrets, so the companies usually reject to share independent and private data sources due to business competition and privacy concerns.} As a result, breaking data silos for multi-agent perception is never studied before.

To address the Distribution Gap issue, we introduce the \textbf{F}eature  \textbf{D}istribution-aware  \textbf{A}ggregation framework for Multi-agent perception system on 3D object detection task, named as FDA. We choose the task of V2X cooperative perception for the point cloud-based 3D object detection as an investigation in this paper.  
Specifically, we proposed two key components: 1) Learnable Feature Compensation Module (LFCM) to generate a residual compensation map for CAV's intermediate feature considering the large-scale feature information; 2) Distribution-aware Statistical Consistency Module (DSCM) to diminish the distribution gap between CAV feature and ego feature in terms of feature distributions. We conducted extensive experiments on two public V2X perception datasets, namely OPV2V~\cite{xu2022opv2v} and V2XSet~\cite{xu2022v2x}, to justify the effectiveness of our proposed method. Our contributions are summarized as follows:

\begin{itemize}
    \item To the best of our knowledge, we propose a novel Feature Distribution-aware Aggregation framework, dubbed as FDA, which is the \textbf{first research} on multi-agent perception to address the Distribution Gap of different independent private data for training distinct agents.      

    \item We propose a novel Learnable Feature Compensation Module (LFCM) to effectively mitigate discrepancies arising from various agents and a Distribution-aware Statistical Consistency Module (DSCM) to diminish the differences between intermediate features extracted from the ego agent and other agents.

    \item We evaluate the proposed FDA framework on the large-scale simulated OPV2V and V2XSet datasets, whose experimental results demonstrate our superior performance in point cloud-based 3D object detection.    
\end{itemize}

\section{Related Work}\label{Sec:Related_Work}

\noindent \textbf{Perception on multi-agent system.}
Multi-vehicle perception systems aim to overcome the limitations of single-vehicle systems by leveraging multi-vehicle information. Collaboration modules are commonly developed by researchers to improve efficiency and performance.
Typically, three approaches are used for aggregating multi-vehicle observations: raw data fusion, feature combination during processing, and output fusion. State-of-the-art methods often opt for sharing intermediate neural features to balance accuracy and bandwidth requirements~\cite{xu2022opv2v,xu2022v2x,wang2020v2vnet}.
Attfuse~\cite{xu2022opv2v} harnesses self-attention models for the decompressed features amalgamation.
SCOPE~\cite{yang2023spatio} introduces a learning-based framework that addresses multi-agent challenges with a first focus on the temporal context of the ego agent.
V2X-ViT~\cite{xu2022v2x} presents a unified ViT architecture for V2X perception which is able to capture the heterogeneous nature of V2X systems.
CoBEVT~\cite{xu2023cobevt} is a pioneering multi-agent perception framework that collaboratively generates predictions using a ViT to enhance performance.
Despite the impressive performance of these methods in V2V perception, their deep learning-based network is trained with homogeneous data from ego vehicles and other vehicles. There exists an inherent data gap in the real world. In this paper, we aim to tackle this challenge.\\
\noindent \textbf{Vehicle-to-Everything Dataset.}
Numerous datasets, \textit{e.g.}, Cityscapes~\cite{cordts2016cityscapes} and KITTI~\cite{geiger2012we}, serve single-agent perception. Large-scale datasets are crucial for robust cooperative perception models in multi-agent systems. However, collecting real-world multi-agent data is challenging and costly.
To address this, simulators like CARLA~\cite{dosovitskiy2017carla} and OpenCDA~\cite{10045043} are used to gather cooperative perception data. V2X-Sim~\cite{li2022v2x}, for instance, provides a simulated dataset for multi-agent perception in V2X-assisted autonomous driving, including records from various agents.
For simulated V2V perception, OPV2V~\cite{xu2022opv2v} offers an extensive open-source dataset with diverse scenes simulating real-world traffic dynamics, including a digital rendition of Culver City.
In real-world data, DAIR-V2X~\cite{yu2022dair} features multi-modal, multi-view data extracted from real-world Vehicle-Infrastructure Cooperative Autonomous Driving scenarios. V2V4Real is exclusively tailored for V2V cooperative autonomous perception.
To address the shortage of real-world sequential data, V2X-Seq~\cite{yu2023v2x} captures essential elements from natural scenery.
Current state-of-the-art cooperative perception methods assume uniform data distributions for ego vehicle and surrounding CAV encoders, using identical training datasets. This paper investigates cooperative perception performance in the context of diverse data distributions.\\
\noindent \textbf{Deployment of Multi-Agent System.}
Compared to single-agent systems, multi-agent systems with V2V communication introduce challenges like communication latency, lossy communication, localization errors, and adversarial attacks, which can undermine collaboration benefits~\cite{xu2023bridging,xu2022v2x}. Recent research has made progress in enhancing system robustness.
V2X-ViT~\cite{xu2022v2x} uses a Vision Transformer (ViT) to handle GPS localization errors and sensing information delays.
To address localization errors, Vadivelu et al.~\cite{vadivelu2021learning} proposed a pose regression module that learns a correction parameter to predict the true relative transformation from noisy data.
The LC-aware Repair Network~\cite{li2023learning} is introduced to enhance collaborative perception robustness under lossy communication conditions, addressing packet loss problems.
A model-agnostic framework~\cite{xu2023model} is proposed to address model heterogeneity in collaborative perception.
~\cite{xu2023bridging} proposes MPDA framework to narrow the domain gap in spatial resolution, channel count, and patterns in multi-agent perception.
~\cite{li2023among} introduces ROBOSAC to enhance adversarially robust collaborative perception by promoting consensus among robots during collaboration.
In this paper, we aim to address the distribution gap arising from differently distributed data to enhance the robustness of cooperative perception.

\section{Methodology}\label{Sec:Method}

\subsection{Overview of the Feature Distribution-aware Aggregation}\label{overview}
The Feature Distribution-aware Aggregation framework via the V2V cooperative perception pipeline is illustrated in Fig.~\ref{fig:FDA}.
First, we select an ego vehicle among the CAVs to create a spatial graph that encompasses nearby CAVs within the communication range. Because of the similar sharing capabilities of CAVs and infrastructures, we consider each intelligent infrastructure as a CAV during methodology development in this paper. All the other nearby CAVs will project their own LiDAR data onto the ego vehicle’s coordinate frame, based on both the ego vehicle’s and their own GPS poses. The point clouds from the ego and CAVs are denoted as $\mathbf{P}_{ego} \in \mathbb{R}^{4 \times m}$ and $\mathbf{P}_{cav} \in \mathbb{R}^{4 \times m}$, respectively.
Each CAV has its own Encoder for LiDAR feature extraction. After feature extraction of each CAV, the ego vehicle receives the neighboring CAV visual features via V2V communication.
The intermediate features aggregated from $N$ surrounding CAVs are denoted as $\mathbf{F}^{cav} \in \mathbb{R}^{N \times H \times W \times C}$, and the ego intermediate features are denoted as $\mathbf{F}^{ego} \in \mathbb{R}^{1 \times H \times W \times C}$. These, along with features from other CAVs, are fed into our proposed LFCM and DSCM modules to reduce feature disparities.
After processing by our proposed modules, the intermediate features are fused using the Feature Fusion Network (FFN).
Finally, the resulting fused feature maps are fed into a prediction header for 3D bounding-box regression and classification.
We formulate the original V2V cooperative perception system for LiDAR-based 3D object detection as
~
\begin{equation}
    \digamma (\mathbf{P}_{cav},\mathbf{P}_{ego}) =  \mathbf{H}(\text{FFN}(\mathbf{F}^{cav}, \mathbf{F}^{ego})),
    \label{eq:v2v_total}
\end{equation}
~
\begin{equation}
    \mathbf{F}^{cav} = \mathbf{E}_{cav}(\mathbf{P}_{cav}), \\
    \ \mathbf{F}^{ego} = \mathbf{E}_{ego}(\mathbf{P}_{ego}),
    \label{eq:ego}
\end{equation}
%
where $\text{FFN}(\cdot)$ is the Feature Fusion Network responsible for fusing the features of CAVs and the ego vehicle, and $\mathbf{H}$ is the prediction header for 3D object detection. Existing cooperative perception methods assume that the encoder weights ($w$) of $\mathbf{E}_{ego}$ and $\mathbf{E}_{cav}$ are shared, indicating they are trained with the same distribution data (\textit{i.e.}, $\mathbf{E}_{ego}^{w} = \mathbf{E}_{cav}^{w} $). However, in the real-world deployment, $\mathbf{E}_{ego}^{w} \neq \mathbf{E}_{cav}^{w} $, leading to a distribution gap between CAVs and the ego vehicle for most existing models. To address this, we propose a feature distribution-aware aggregation framework, which can serve as a plug-in module in the V2V cooperative perception system. Then, the Eq.~(\ref{eq:v2v_total}) can be modified as 
\begin{align}
    \digamma (\mathbf{P}_{cav},\mathbf{P}_{ego}) =  \mathbf{H}(\text{FFN}(\text{FDA}(\mathbf{F}_{cav},\mathbf{F}_{ego}), \mathbf{F}_{ego})),
    \label{eq:fda_total}
\end{align}
where $\text{FDA}(\cdot)$ refers to our proposed Feature Distribution-aware Aggregation framework, comprising the learnable feature compensation module and the distribution-aware statistical consistency module, as depicted in Fig.~\ref{fig:FDA}.

\subsection{Learnable Feature Compensation Module}
\label{LFRM}
After encoding by $\mathbf{E}_{ego}$ and $\mathbf{E}_{cav}$, we obtain intermediate neural features $\mathbf{F}^{ego}$ and $\mathbf{F}^{cav}$ from the ego vehicle and other CAVs, respectively.
Due to the difference between $\mathbf{E}_{ego}^{w}$ and $\mathbf{E}_{cav}^{w}$ based on distinct training data, even when the same point cloud is fed into both encoders, different feature maps are obtained. To mitigate the impact of varying weights in the encoders, we introduce a Learnable Feature Compensation Module (LFCM). The LFCM is an encoder-decoder architecture with skip connections, inspired by the work of~\cite{mildenhall2018burst}, and depicted in Fig.~\ref{fig:FDA}. The encoder and decoder components of our LFCM employ $5 \times 5$ convolutional layers to capture large-scale spatial features.
The shared features $\mathbf{F}^{cav}$ from CAVs are fed into the LFCM to predict residual compensation scores for the entire feature set, generating a spatial compensation map $\mathbf{M}_{c} \in \mathbb{R}^{N \times H \times W \times C}$ of the same size as $\mathbf{F}^{cav}$. Subsequently, $\mathbf{M}_{c}$ is added to $\mathbf{F}^{cav}$, incorporating residual compensation levels into consideration, resulting in the enhanced compensated feature $\mathbf{\widehat{F}}^{cav}$.
The computation of the proposed LFCM is formulated as
\begin{align}
    \mathbf{\widehat{F}}^{cav} = \mathrm{LFCM}(\mathbf{F}^{cav}) + \mathbf{F}^{cav},
    \label{eq:lfcm}
\end{align}
where $\mathrm{LFCM} (\cdot)$ represents our proposed LFCM responsible for generating the residual compensation map.

\subsection{Distribution-aware Statistical Consistency Module }
\label{DSCM}
After processing with our LFCM, we obtain enhanced compensated features for the CAV, denoted as $\mathbf{\widehat{F}}^{cav}$, and the ego intermediate features $\mathbf{F}^{ego}$.
To address the distribution gap between $\mathbf{\widehat{F}}^{cav}$ and $\mathbf{F}^{ego}$, it is crucial to examine the relationship between statistical differences and feature distributions. Previous studies~\cite{hou2023evading} have established a positive correlation between statistical differences and distribution disparities.
In order to minimize the discrepancy in feature distributions between $\mathbf{\widehat{F}}^{cav}$ and $\mathbf{F}^{ego}$, we introduce the Maximum Mean Discrepancy (MMD)~\cite{JMLR:v13:gretton12a} distance as a metric. Let $\mathcal{F}^{ego} = \{\mathbf{F}_{i}^{ego}\}$ and $\mathcal{F}^{cav} = \{\mathbf{\widehat{F}}_{i}^{cav}\}$ represent the sets of ego and compensated CAV features, respectively.
Our goal is to enhance the compensated CAV features to reduce the distribution gap with ego features in terms of feature distributions by tuning the parameters of $\mathrm{LFCM}$. This process can be formulated as follow:
\begin{equation}
    \mathrm{arg} \min\limits_{\mathrm{LFCM}} L_{mmd}(\mathcal{F}^{ego}, \mathcal{F}^{cav} ),
    \label{eq:mmd}
\end{equation}
where $L_{mmd}(\cdot)$ represents the MMD loss between the two sets of intermediate features~\cite{JMLR:v13:gretton12a}.

\subsection{Loss Function}
\label{loss}
For 3D object detection, as in~\cite{xu2023bridging, xu2022v2x}, we compute the smooth L1 loss for bounding box regression and apply focal loss~\cite{lin2017focal} for classification.
In the context of our FDA framework, we employ the Maximum Mean Discrepancy (MMD) loss to address distribution differences between ego and other CAV agents.
The final loss function is a combination of these two losses:
\begin{equation}
    L_{total} = \lambda L_{det} + \omega L_{mmd},
    \label{eq:loss}
\end{equation}
where $\lambda$ and $\omega$ are the balancing coefficients, both ranging within [0, 1].
\begin{table*}[htb]
\caption{\textbf{3D detection performance on OPV2V testing set under \textit{V2XSet $\rightarrow$ OPV2V} setting}. We show Average Precision (AP) at IoU=0.5, 0.7. \textit{Dist. Gap} represents the distribution gap scenario, \textit{No Dist} represents agents' encoders consistently trained on the same training set. \textit{Dist} stands for Agents' encoders trained on different training sets, causing a distribution gap. Note that the training set type for original CAV encoders and ego encoder training is defined as $S_{cav}$, $S_{ego}$, respectively.}
\centering
\label{tab:opv2v}
\resizebox{1\textwidth}{!}{%
\begin{tabular}{@{}cc|cc|cc|cc|cc|cc@{}}
\toprule
\multirow{2}{*}{$S_{cav} \Vert \ S_{ego}$} &
  \multirow{2}{*}{\textit{Dist. Gap}} &
  \multicolumn{2}{c|}{NO Fusion} &
  \multicolumn{2}{c|}{Attfuse~\cite{xu2022opv2v}} &
  \multicolumn{2}{c|}{V2VAM~\cite{10077757}} &
  \multicolumn{2}{c|}{V2X-ViT~\cite{xu2022v2x}} &
  \multicolumn{2}{c}{CoBEVT~\cite{xu2023cobevt}} \\
                           &             & AP@0.5 & AP@0.7 & AP@0.5 & AP@0.7 & AP@0.5 & AP@0.7 & AP@0.5 & AP@0.7 & AP@0.5 & AP@0.7 \\ \midrule
OPV2V $\Vert$  OPV2V    &\textit{No Dist}  & 67.9  & 60.2  & 90.7  & 81.0   & 91.6  & 84.9  & 90.2  & 79.2  & 92.5  & 85.2  \\
V2XSet $\Vert$  OPV2V   & \textit{Dist}   & 67.9  & 60.2  & 36.1  & 13.0   & 74.7  & 56.6  & 42.7  & 19.5  & 72.7  & 55.2  \\
V2XSet $\Vert$  OPV2V   & \textit{Dist} (Finetuned) & 67.9  & 60.2  & 69.6  & 43.4  & 87.7  & 72.0   & 74.7  & 36.6  & 77.0   & 56.0   \\
V2XSet $\Vert$  OPV2V   & \textit{Dist} + FDA        & 67.9  & 60.2  & \textbf{85.7}  & \textbf{69.0}   & \textbf{88.3}    & \textbf{79.4}    & \textbf{89.7}    & \textbf{75.2}    & \textbf{89.0}   & \textbf{74.7}  \\ \bottomrule
\end{tabular}%
}
\end{table*}

\begin{table*}[htb]
\caption{\textbf{3D detection performance on V2XSet testing set under \textit{OPV2V $\rightarrow$ V2XSet} setting}. We show Average Precision (AP) at IoU=0.5, 0.7. Note that the training set type for original CAV encoders and ego encoder training  are defined as $S_{cav}$, $S_{ego}$, respectively}
\centering
\label{tab:v2xset}
\resizebox{1\textwidth}{!}{%
\begin{tabular}{@{}cc|cc|cc|cc|cc|cc@{}}
\toprule
\multirow{2}{*}{$S_{cav} \Vert \ S_{ego}$} &
  \multirow{2}{*}{\textit{Dist. Gap}} &
  \multicolumn{2}{c|}{NO Fusion} &
  \multicolumn{2}{c|}{Attfuse~\cite{xu2022opv2v}} &
  \multicolumn{2}{c|}{V2VAM~\cite{10077757}} &
  \multicolumn{2}{c|}{V2X-ViT~\cite{xu2022v2x}} &
  \multicolumn{2}{c}{CoBEVT~\cite{xu2023cobevt}} \\
                           &             & AP@0.5 & AP@0.7 & AP@0.5 & AP@0.7 & AP@0.5 & AP@0.7 & AP@0.5 & AP@0.7 & AP@0.5 & AP@0.7 \\ \midrule
V2XSet $\Vert$ V2XSet    &\textit{No Dist}  & 60.6  & 40.2  & 79.2  & 64.1   & 90.2  & 81.6  & 88.2  & 71.2  & 84.4  & 66.0  \\
OPV2V $\Vert$  V2XSet   & \textit{Dist}   & 60.6  & 40.2  & 45.1  & 32.7   & 70.6  & 53.9  & 50.0  & 36.4  & 55.5  & 40.7  \\
OPV2V $\Vert$  V2XSet   & \textit{Dist} (Finetuned)& 60.6  & 40.2  & 64.6  & 46.8  & 84.6  & 70.4   & 68.0  & 47.9  & 74.9   & 54.7\\
OPV2V $\Vert$  V2XSet   & \textit{Dist} + FDA        & 60.6  & 40.2  & \textbf{77.2}  & \textbf{57.2}   & \textbf{89.7}    & \textbf{81.3}    & \textbf{79.0}    & \textbf{59.6}    & \textbf{82.9}   & \textbf{64.3}  \\ \bottomrule
\end{tabular}%
}
\end{table*}

\begin{figure*}[!t]
\centering
\subfloat[V2X-ViT with \textit{Dist} ]{
  \includegraphics[width=0.64\columnwidth]{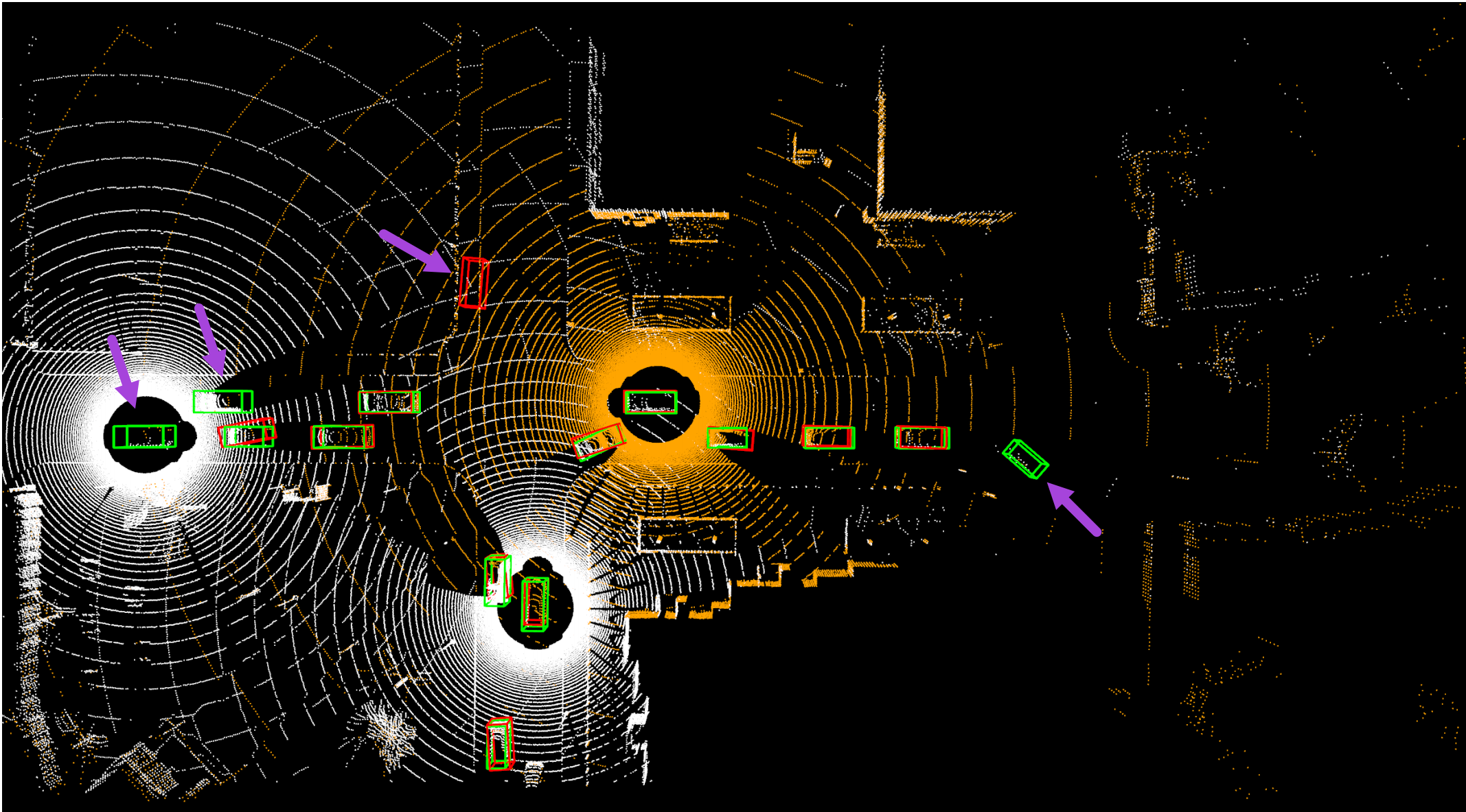}
}
\subfloat[ Finetuned V2X-ViT with \textit{Dist}]{
  \includegraphics[width=0.64\columnwidth]{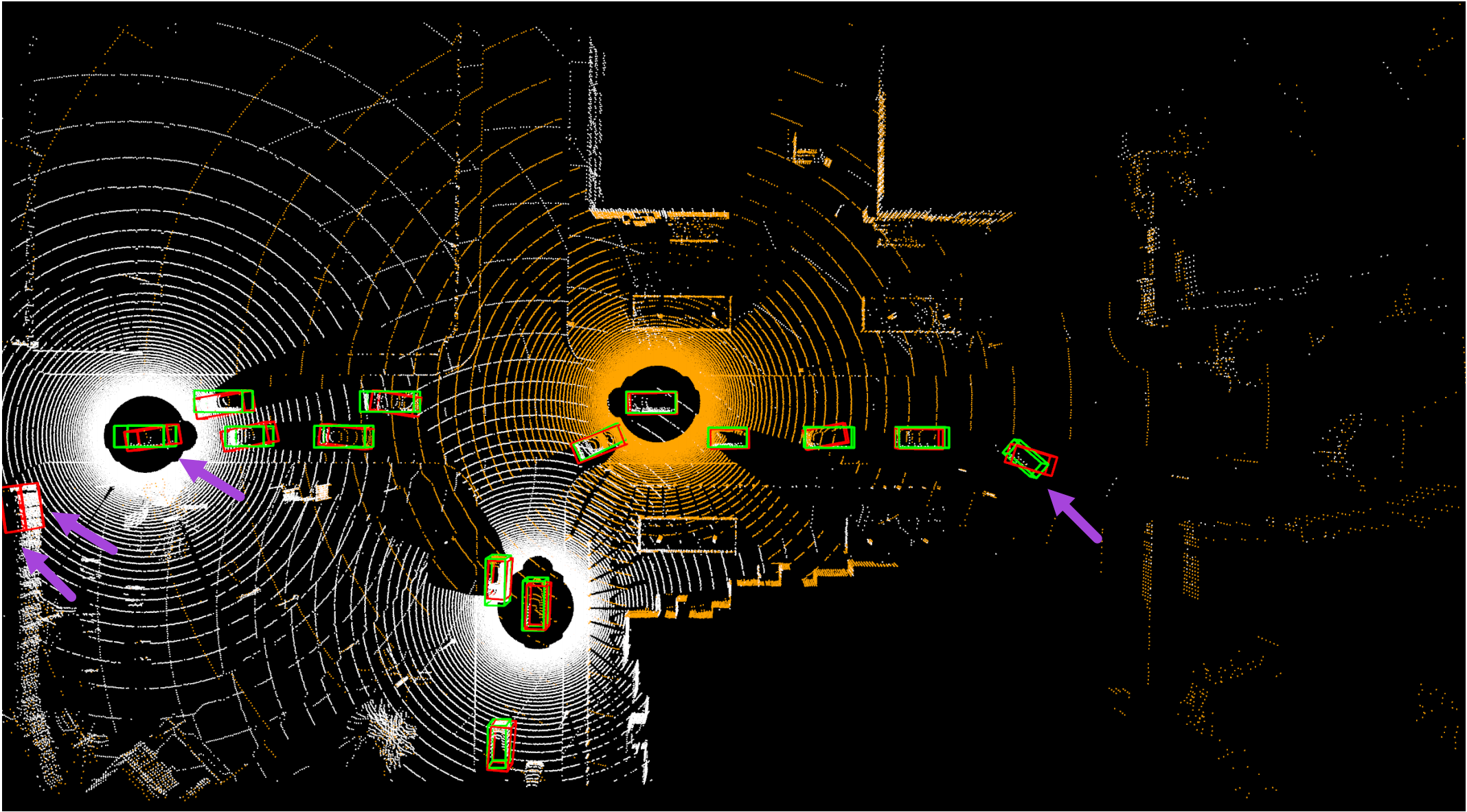}
}
\subfloat[V2X-ViT + FDA with \textit{Dist}]{
  \includegraphics[width=0.64\columnwidth]{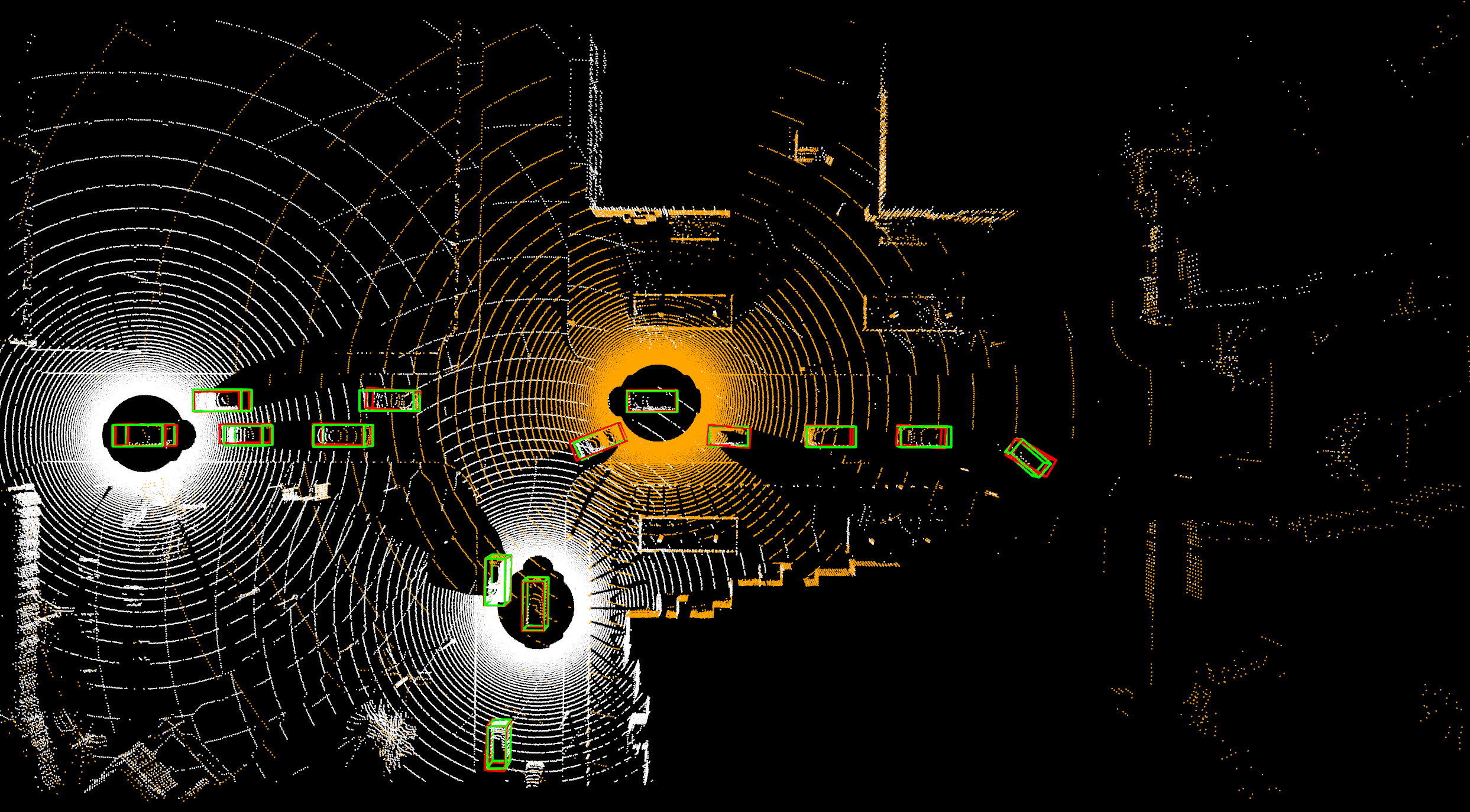}
}\\
\subfloat[CoBEVT with \textit{Dist}]{
  \includegraphics[width=0.64\columnwidth]{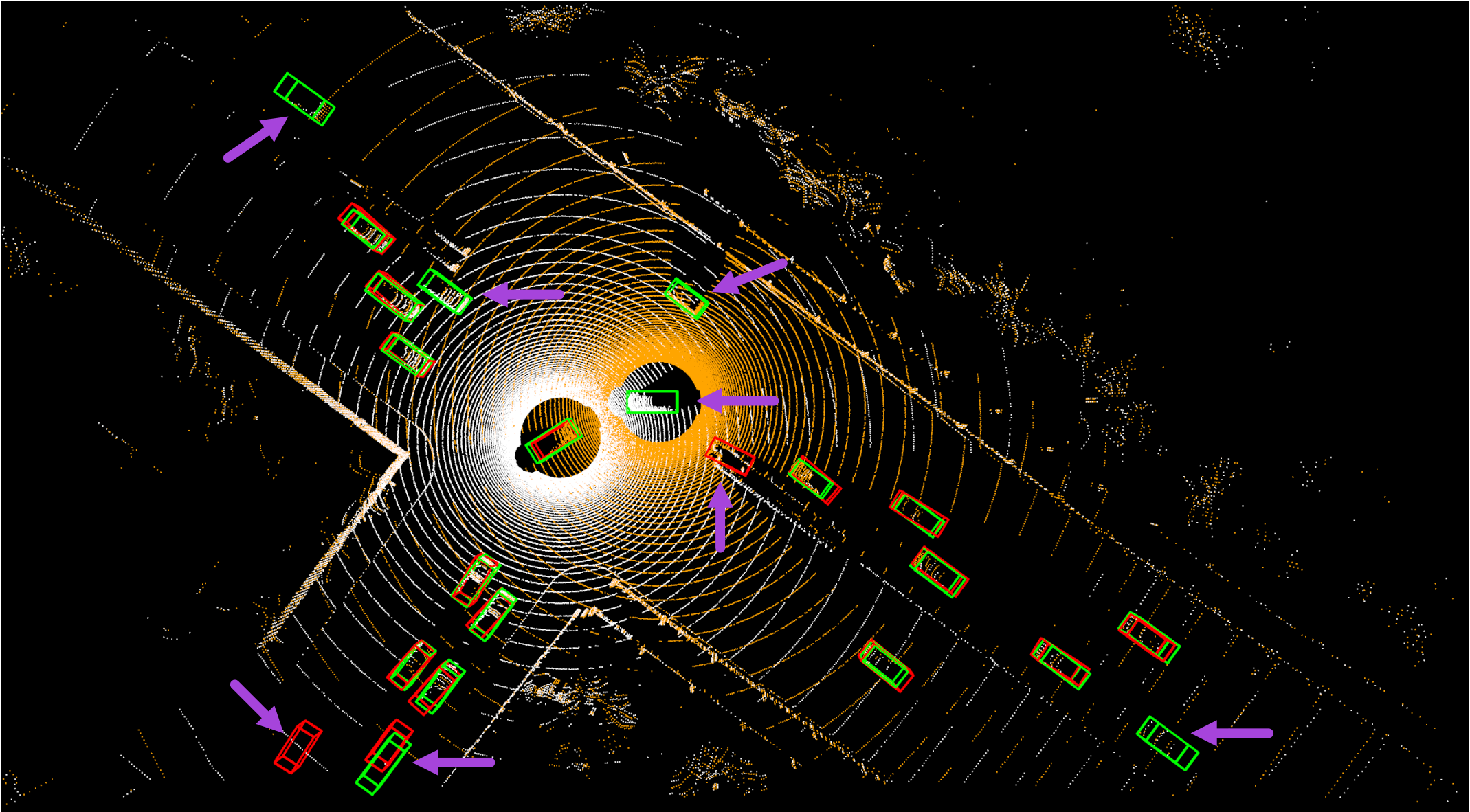}
}
\subfloat[Finetuned CoBEVT with \textit{Dist}]{
  \includegraphics[width=0.64\columnwidth]{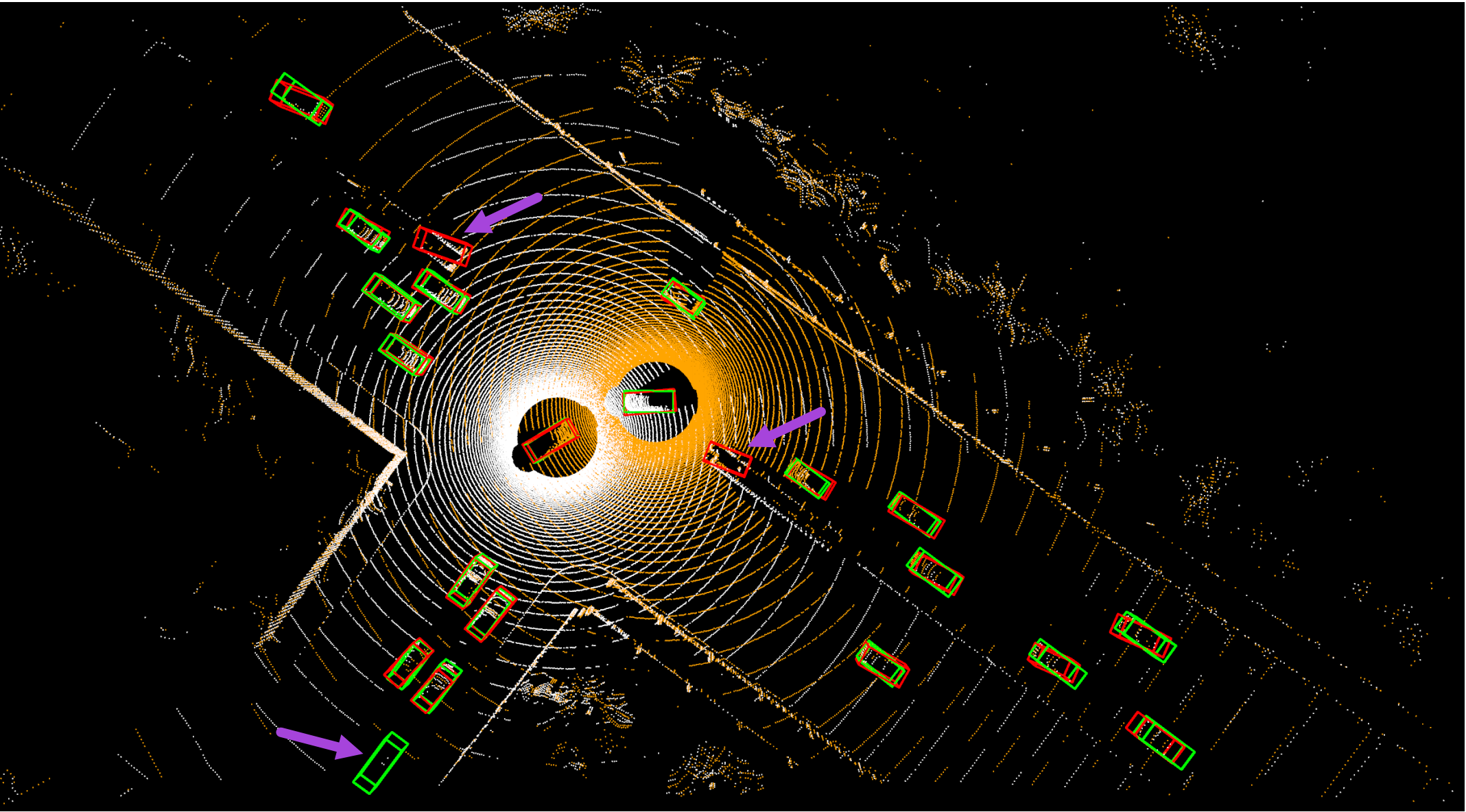}
}
\subfloat[CoBEVT + FDA  with \textit{Dist} ]{
  \includegraphics[width=0.64\columnwidth]{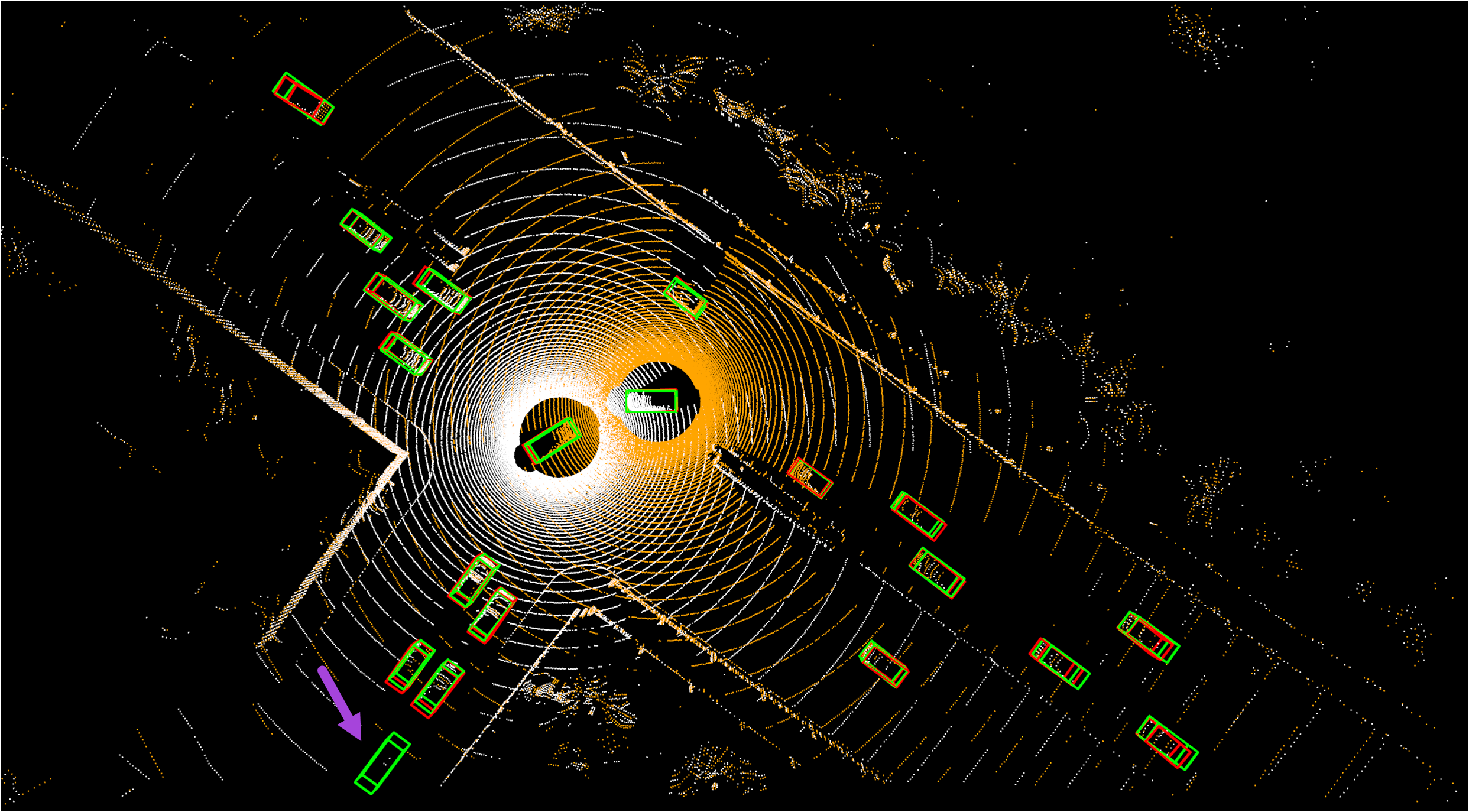}
}
\caption{\textbf{3D object detection visualization.} {\color{amber(sae/ece)} Orange} point cloud is ego vehicle, and the white color of point clouds are CAVs. \textcolor{green}{Green} and \textcolor{red}{red} 3D bounding boxes represent the \textcolor{green}{ground truth} and \textcolor{red}{prediction} respectively. The detection results of the proposed FDA are clearly more accurate. False detection errors are highlighted using purple arrows.}
\label{fig:visualization}
\end{figure*}

\begin{figure*}[]
    \begin{centering}
        \includegraphics[width=2\columnwidth]{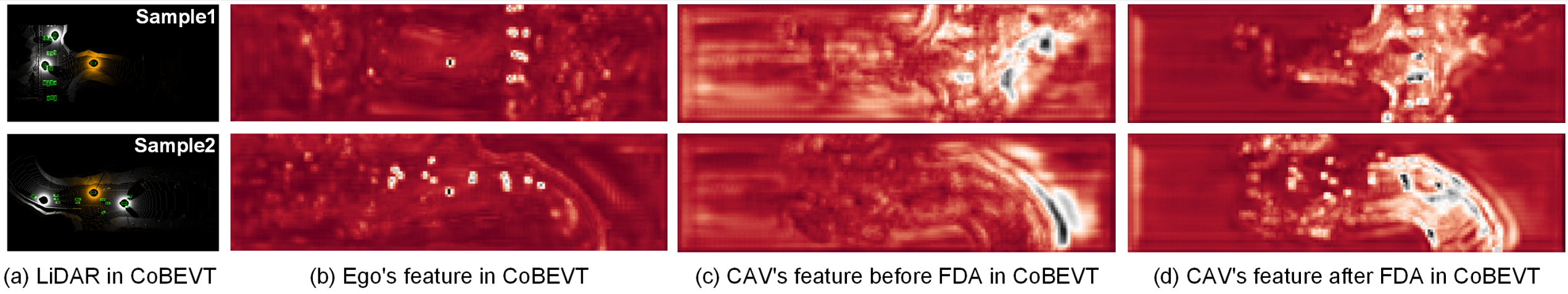}
        \par\end{centering}
    \caption{\textbf{Visualization of intermediate features before and after our FDA.} Two samples of point clouds are selected on OPV2V testing set to evaluate  CoBEVT~\cite{xu2023cobevt}, where {\color{amber(sae/ece)} Orange} point cloud is ego vehicle. It is evident that after applying the FDA, the intermediate features from CAVs exhibit more similar patterns to those of the ego. Bright pixels may tend to represent the objects.}
    \label{fig:featureMap}
\end{figure*}

\section{Experiments}\label{Sec:Experiment}

\noindent \textbf{Dataset}: Our experiments are conducted on two publicly available benchmark datasets for V2V/V2X cooperative perception tasks: OPV2V~\cite{xu2022opv2v} and V2XSet~\cite{xu2022v2x}.
\textbf{OPV2V} is a large-scale simulated dataset tailored for V2V cooperative perception tasks, collected using the CARLA~\cite{dosovitskiy2017carla} and OpenCDA~\cite{10045043} platforms. It comprises 73 diverse scenes featuring varying numbers of connected vehicles. The dataset is divided into training (6,764 frames), validation (1,981 frames), and testing (2,719 frames) sets.
\textbf{V2XSet} is another extensive simulated dataset designed for V2X cooperative perception tasks, also collected using CARLA. This dataset provides LiDAR data from multiple autonomous vehicles and roadside intelligent infrastructure, all timestamped within the same scenarios. Its training/validation/testing set is split into 6,694, 1,920, and 2,833 frames, respectively

\subsection{Experimental Setup}
\noindent \textbf{Evaluation Metrics:} 
Following~\cite{xu2022opv2v,xu2022v2x}, our performance evaluation centers around the final 3D vehicle detection accuracy. We set the evaluation range as $x\in[-140, 140]$ meters,  $y\in[-40, 40]$ meters, where all CAVs are included in this spatial range in the experiment. We measure the accuracy with Average Precisions (AP) at Intersection-over-Union (IoU) thresholds of $0.5$ and $0.7$.

\noindent \textbf{Experimental Details:}
In our investigation aimed at addressing the distribution gap in LiDAR-based 3D object detection, we make the assumption that all agents employ the same cooperative perception method. However, there is a distinction between the weight of encoders used for the ego vehicle and the CAVs, denoted as $\mathbf{E}_{ego}^{w} \neq \mathbf{E}_{cav}^{w}$. Unlike traditional training strategies where all encoders are trained on the same dataset, we introduce a novel approach.
Specifically, we use the OPV2V and V2XSet training sets to train the ego encoder and CAV encoders, respectively. This configuration involves scenarios where the CAV encoders are trained on the OPV2V training set, while the ego encoder is trained on the V2XSet training set, defined as \textit{OPV2V $\rightarrow$ V2XSet}. The testing sets of these datasets (\textit{i.e.}, OPV2V and V2XSet) are used to evaluate our cooperative methods.
We assess our models under two key settings:
\begin{enumerate}
    \item  \textit{\textbf{V2XSet $\rightarrow$ OPV2V:}} All cooperative perception methods are evaluated on the OPV2V testing set, with CAV encoders trained on V2XSet training set while the ego encoder is trained on OPV2V training set.
    \item \textit{\textbf{OPV2V $\rightarrow$ V2XSet:}} Here, all cooperative perception methods are evaluated on the V2XSet testing set, with CAV encoders trained on OPV2V training set while the ego encoder is trained on V2XSet training set.
\end{enumerate}

\noindent \textbf{Compared Methods:}
We utilize four state-of-the-art intermediate fusion methods to assess the distribution gap, comprising two attention-based fusion methods, Attfuse~\cite{xu2022opv2v} and V2VAM~\cite{10077757}, and two ViT-based fusion methods, V2X-ViT~\cite{xu2022v2x} and CoBEVT~\cite{xu2023cobevt}. We establish a \textit{No Fusion} baseline where no collaboration is involved in the system. To illustrate the significant impact of the distribution gap, we train and evaluate these four fusion methods under two distinct settings. Additionally, we finetune these models based on the specific dataset used for the ego encoder.
Furthermore, to showcase the effectiveness of our proposed FDA framework in addressing the distribution gap, we apply our proposed LFCM and DSCM to finetune these fusion methods. This fintuning process serves to mitigate the distribution gap between the ego and CAV encoders.

\noindent \textbf{Implementation Details:}
All cooperative perception models utilize PointPillar~\cite{lang2019pointpillars} as the backbone. We employ the Adam optimizer~\cite{loshchilov2017decoupled} with an initial learning rate of $10^{-3}$, gradually decaying it every 10 epochs using a factor of 0.1. The hyperparameters align with those settings of~\cite{xu2022v2x}. To simulate real-world scenarios where CAV encoders originate from diverse sources that are challenging to update, we freeze the parameters of CAV encoders during finetuning. All models are trained using two RTX 3090 GPUs. The coefficients for the detection loss ($\lambda$) and our MMD loss ($\omega$) are both set to 1.0.

\begin{table}[]
\centering
\caption{Ablation study of the proposed FDA method (LFCM + DSCM) on OPV2V testing set under \textit{V2XSet $\rightarrow$ OPV2V} setting. We show Average Precision (AP) at IoU=0.5.}
\tiny
\centering
\resizebox{1.0\columnwidth}{!}{%
\begin{tabular}{cccc}
\toprule
LFCM &  DSCM & V2X-ViT~\cite{xu2022v2x} & CoBEVT~\cite{xu2023cobevt} \\ \midrule

           &             & 42.7 (\textit{Dist})           & 72.7 (\textit{Dist})      \\
\checkmark &             & 79.1 (\textbf{\color{gray} +36.4})    & 80.3 (\textbf{\color{gray} +7.6})       \\
           & \checkmark  & 83.2 (\textbf{\color{gray} +40.5})            &  83.9 (\textbf{\color{gray} +11.2})        \\
\checkmark & \checkmark  & 89.7 (\textbf{\color{gray} +47.0})            & 89.0 (\textbf{\color{gray} +16.3})   \\ \bottomrule
\end{tabular}
    }
\label{tab:abla}
\end{table}

\subsection{Quantitative Evaluation}

\noindent\textbf{Performance Analysis of Distribution Gap:}
Table~\ref{tab:opv2v} provides the 3D object detection results on the OPV2V testing set. 
Under the  \textit{V2XSet $\rightarrow$ OPV2V} setting, we explore the impact of distribution disparities on the detection performance.
Compared to the \textit{No Dist}, where all agents are consistently trained with the OPV2V training set, the \textit{Dist} for Attfuse~\cite{xu2022opv2v} and V2X-ViT~\cite{xu2022v2x} exhibits a significant drop to $36.1\%/13.0\%$ and $42.7\%/19.5\%$ for AP@0.5/0.7, respectively.
In the \textit{Dist}, the agents of CAVs are trained on the V2XSet training set, leading to even lower performance than a single-agent perception system, \textit{i.e.}, \textit{NO Fusion}. These notable performance drops underscore the highly negative impact of the distribution gap.
We perform finetuning on the four intermediate fusion methods using the OPV2V training set, specifically enhancing the ego encoder and feature fusion network while keeping the CAVs' encoder parameters fixed.
Finetuning improves performance in all fusion methods, but they still fall short of the \textit{NO Fusion} baseline in AP@0.7.
The key reason behind this limitation is that the CAVs' encoder parameters remain fixed, potentially leading to suboptimal feature extraction from the new and previously unseen distribution data.
Our FDA method effectively addresses these limitations. When applied to all fusion methods, FDA achieves substantial improvements that nearly restore the original detection performance under a distribution gap. For instance, Attfuse and V2X-ViT with FDA exhibit impressive improvements of $49.6\%/56.0\%$ and $47.0\%/55.7\%$ for AP@0.5/0.7, respectively.
Furthermore, the results presented in Table~\ref{tab:v2xset} also demonstrate the effectiveness of our proposed FDA method on the V2XSet dataset. Clearly, our proposed FDA method efficiently preserves the collaborative benefits under different distribution-based agents. It successfully mitigates the impact of distribution gaps, resulting in excellent cooperative perception performance.\\
\noindent \textbf{Ablation Study:}
Table~\ref{tab:abla} presents the results of using two specially designed components in FDA framework on the detection performance of V2X-ViT~\cite{xu2022v2x} and CoBVET~\cite{xu2023cobevt}.
The incorporation of LFCM and DSCM into V2X-ViT~\cite{xu2022v2x} leads to performance improvements of $36.4\%$ and $40.5\%$, respectively, showing the effectiveness of our design.\\
\noindent\textbf{3D Detection Visualization:}
We present visual comparisons of various methods in the OPV2V testing set scenario, illustrating their impact on V2X-ViT~\cite{xu2022v2x} and CoBEVT~\cite{xu2023cobevt} in Fig.~\ref{fig:visualization}.
Under the \textit{V2XSet $\rightarrow$ OPV2V} setting, V2X-ViT and CoBEVT exhibit numerous false-negative and false-positive detection errors when no specially designed modules are applied.
Even after direct finetuning of these two models on the OPV2V training set, detection performance improves, but noticeable missing detections and false positives persist, as evident in (b) and (e) of Fig.~\ref{fig:visualization}.
However, when employing our proposed FDA, which takes into account the distribution gap, V2X-ViT and CoBEVT demonstrate more robust performance, successfully detecting most objects and mitigating false-negative and false-positive detection errors.\\
\noindent\textbf{Feature Visualization:}
To analyze the impact of the distribution gap on agents, we visually present intermediate features in Fig.~\ref{fig:featureMap} using two point cloud samples. 
Under the \textit{V2XSet $\rightarrow$ OPV2V} setting, observable disparities exist between ego and CAVs' features. Our FDA effectively alleviates this gap, rendering the features more akin to those of the ego. In the case of CoBEVT, our FDA enhances local details, narrowing the gap with ego's features, as seen in (b-d). This visual evidence underscores the effectiveness of our FDA.

\section{Conclusions}\label{Sec:Conclusions}

In this paper, we present pioneering research on the Distribution Gap, stemming from disparate private training data for distinct agents in multi-agent perception systems, resulting in data silos. We analyze the impact of the Distribution Gap on existing cooperative perception methods. To mitigate data silos, we propose a Feature Distribution-aware Aggregation framework, comprising a Learnable Feature Compensation Module and a Distribution-aware Statistical Consistency Module. We empirically assess the FDA framework on two public and  extensive simulated datasets, OPV2V and V2XSet, demonstrating its superior performance in point cloud-based 3D object detection. Our findings emphasize the crucial role of addressing the distribution gap in multi-agent perception systems, contributing to the advancement of cooperative perception strategies with potential implications for autonomous driving.

\bibliographystyle{IEEEtran}
\bibliography{Jinlong}

\end{document}